\documentclass{Interspeech}

\interspeechcameraready 

\usepackage{cite}
\usepackage{microtype}

\title{Spoken Language Modeling with Duration-Penalized Self-Supervised Units}

\author[]{Nicol}{Visser}
\author[]{Herman}{Kamper}

\affiliation[nocounter]{Electrical and Electronic Engineering}{Stellenbosch University}{South Africa}
\email{16986431@sun.ac.za, kamperh@sun.ac.za}
\keywords{spoken language modeling, acoustic unit discovery, discrete speech units, duration-penalized units}

\usepackage{bbm} 
\usepackage{tabularx}
\usepackage{multirow}

\newcolumntype{C}{>{\centering\arraybackslash}X}
\newcolumntype{L}{>{\raggedright\arraybackslash}X}
\newcolumntype{R}{>{\raggedleft\arraybackslash}X}

\begin{document}

\maketitle

\begin{abstract}
Spoken language models (SLMs) operate on acoustic units obtained by discretizing self-supervised speech representations. Although the characteristics of these units directly affect performance, the interaction between codebook size and unit coarseness (i.e., duration) remains unexplored. We investigate SLM performance as we vary codebook size and unit coarseness using the simple duration-penalized dynamic programming (DPDP) method. New analyses are performed across different linguistic levels. At the phone and word levels, coarseness provides little benefit, as long as the codebook size is chosen appropriately. However, when producing whole sentences in a resynthesis task, SLMs perform better with coarser units. In lexical and syntactic language modeling tasks, coarser units also give higher accuracies at lower bitrates. We therefore show that coarser units aren't always better, but that DPDP is a simple and efficient way to obtain coarser units for the tasks where they are beneficial.
\end{abstract}

\section{Introduction}

Spoken language models (SLMs) aim to learn language directly from speech audio~\cite{dunbar2021zrc}.
In contrast to models that combine speech and text\cite{nguyen2024spiritlm, hassid2024twist}, SLMs do not use text data in any of their components.
Although this could enable speech technology in settings where text cannot be collected, SLMs still lag far behind text language models~\cite{poli2023slmfinetunetext}.
One potential reason lies in the nature of the discrete speech units that SLMs use as input.
These units, typically obtained by discretizing self-supervised speech representations \cite{lakhotia2021gslm, borsos2023audiolm}, are often much shorter than categorical linguistic units like phones~\cite{algayres2023wordsized, kamper2020vqwordseg, bhati2022segmentationCPC}.
Transformer-based SLMs, therefore, need to deal with much longer input sequences than text language models.

A potential solution is to use methods that produce coarser units spanning longer durations.
Prior work has explored using automatically discovered syllable-like~\cite{baade2024syllablelm} and word-like~\cite{algayres2023wordsized} units as inputs to SLMs.
However, these methods either require complex modeling techniques~\cite{baade2024syllablelm}, or offer only marginal improvements over finer-grained units~\cite{algayres2023wordsized}.
A more straightforward method is to use byte-pair encoding to find recurring patterns within the quantized unit sequences and use these patterns as SLM units.
However, the reported benefits of this approach remain inconsistent across different studies \cite{shen2024acousticbpe, baade2024syllablelm}.

Another factor that indirectly influences coarseness is the codebook size used for quantizing the self-supervised speech features.
Previous work~\cite{lakhotia2021gslm, borsos2023audiolm} has examined SLM performance across different codebook sizes, indicating that the optimal choice depends on the specific self-supervised model and layer.
Codebook size directly impacts unit expressiveness and, similar to coarseness, affects the bitrate of the resulting unit sequences.

While coarseness and codebook size have been studied in isolation, the interplay between them has not.
As both affect bitrate and interact with each other, their combined effect is a crucial factor in SLM design.
In this paper, we systematically investigate the performance of SLMs as we vary codebook size and unit coarseness.
We employ duration-penalized dynamic programming (DPDP)~\cite{kamper2020vqwordseg, kamper2023dpdp}, a simple and efficient method for creating coarser units.
DPDP uses a cost function to encourage contiguous features to map to the same discrete code in a codebook, resulting in longer units.
Using a uniquely designed SLM framework, we conduct evaluations across a range of linguistic levels, including phone and word discrimination, whole-sentence resynthesis, and lexical and syntactic language modeling.

Our results reveal a nuanced picture. 
At the phone and word levels, increasing unit coarseness provides little benefit.
For these tasks, an appropriately chosen codebook size is often sufficient.
However, for whole-sentence resynthesis and in lexical and syntactic language modeling, there are improvements with coarser units, particularly at higher codebook sizes.
This shows that the benefits of coarser units are not always clear-cut and that the optimal unit granularity depends on the task at hand.
Nevertheless, for scenarios where coarser units are advantageous, DPDP offers a practical and efficient method to obtain them.
Code available at: {\small\href{https://github.com/nicolvisser/dp-slm}{\texttt{https://github.com/nicolvisser/dp-slm}}}

\section{Model and methodology}

An SLM starts by tokenizing a raw waveform into a sequence of discrete units.
This typically involves quantizing framewise features from a self-supervised learning (SSL) model such as \mbox{HuBERT} \cite{hsu2021hubert} or \mbox{WavLM} \cite{chen2021wavlm} and removing consecutive repeated codes (deduplication) to obtain varying-length units.
These units are seen as pseudo-text, enabling the training of a language model that operates directly on audio.
The model can then be probed through tasks like assessing whether the spoken word ``manufacture'' is more probable than ``manufelture.''

Our goal is to see how codebook size and unit coarseness affect SLM performance across various tasks.
To enable this investigation, we use duration-penalized units for tokenization.

\subsection{Duration-penalized dynamic programming (DPDP)}

\begin{figure}[t]
  \centering
  \includegraphics[width=0.99\linewidth]{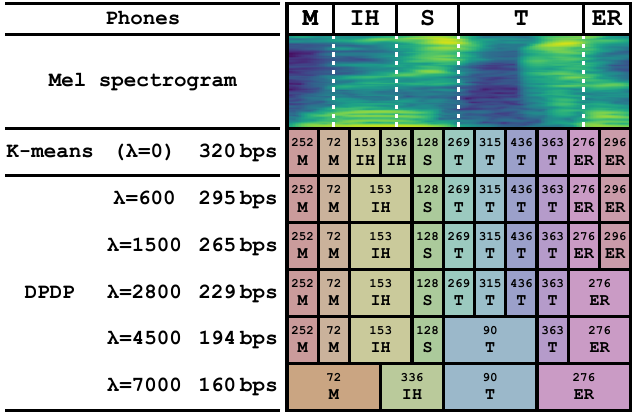}
  \vspace*{-2.5pt}
  \caption{Coarser and coarser speech units obtained with DPDP. The units are labeled with the most likely phone from the forced alignments. The codebook size is $K=500$.}
  \label{fig:dp_units}
\end{figure}

While quantizing speech at a fixed interval can produce variable-length segments if we use deduplication, the units might still not match the duration of real phones.
Duration-penalized dynamic programming (DPDP) is a segmentation method that finds longer units~\cite{kamper2020vqwordseg, kamper2023dpdp}.
DPDP provides a simple way to control unit coarseness while keeping the codebook (or vocabulary) size fixed.
When quantizing SSL features using $K$-means, DPDP encourages contiguous features to be assigned to the same code.

Suppose we have a sequence of SSL features $\mathbf{x}_1,\mathbf{x}_2,\dots,\mathbf{x}_T$ that we want to quantize to a set of $K$ codebook vectors $\{\mathbf{c}_k\}_{k=1}^K$, with each $\mathbf{x}_t, \mathbf{c}_k \in \mathbb{R}^D$.
We want a sequence of code assignments $u_1,u_2,\dots,u_T$, where $u_t \in \{ 1, \dots, K \}$.
DPDP uses dynamic programming to find the sequence that satisfies:\footnote{This formulation is different but exactly equivalent to the one in the original DPDP paper~\cite{kamper2023dpdp}, where the duration term is phrased as a \textit{penalty} rather than a \textit{reward}, as we do here.}
\begin{equation}
    \label{eq:dpdp_objective}
    \min_{u_1,u_2,\dots,u_T} \sum_{t=1}^{T}{ \left( || \mathbf{x}_t - \mathbf{c}_{u_t} ||^2 - \lambda \mathbbm{1}_{u_t = u_{t-1}}  \right)  }
\end{equation}
This cost consists of a quantization penalty $||\mathbf{x}_t-\mathbf{c}_{u_t}||^2$ and a duration reward term $-\lambda \mathbbm{1}_{u_t = u_{t-1}}$ at each timestep.
The duration reward reduces the cost if the proposed code at timestep $t$ is the same as the code at the previous timestep, i.e., longer segments are encouraged.
When no duration reward is given, $\lambda = 0$, the cost in~(\ref{eq:dpdp_objective}) becomes the standard $K$-means cost.
When $\lambda$ is increased, the code assignments may deviate from the codebook entries closest to the SSL features in favor of contiguous assignments to the same code.
This results in a coarser discrete representation.

Figure \ref{fig:dp_units} visualizes the resulting units using progressively higher values for $\lambda$.
We label each unit with the most likely phone from the forced alignments.
As $\lambda$ increases, units corresponding to the same phone merge into larger units, giving a shorter sequence length.
When $\lambda=7000$, the phone labels suggest that the /S/ phone disappears.
However, units 153 (/IH/) and 128 (/S/) have merged into a new unit 336, which previously occurred on the boundary of /IH/ and /S/.
There hasn't been a study on using these larger DPDP units in SLMs.

\subsection{Spoken lanugage model}

\begin{figure}[t]
  \centering
  \includegraphics[width=0.95\linewidth]{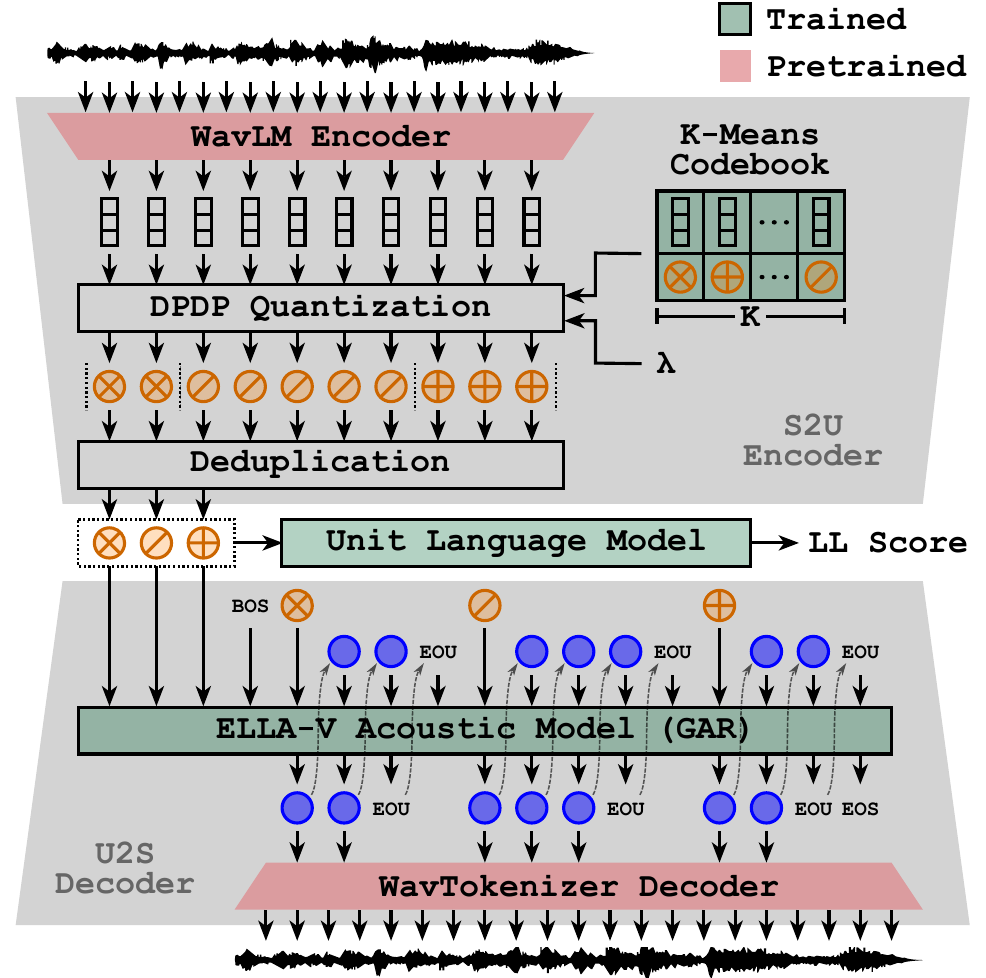}
  \vspace*{-2.5pt}
  \caption{ The spoken language model used in our experiments. The S2U encoder converts speech to discrete units, the unit language model is trained to predict the next unit, and the U2S decoder generates speech from the units.
    }
  \label{fig:pipeline}
\end{figure}

We develop a unique SLM framework for investigating the effect of changing unit coarseness and codebook size, as shown in Figure~\ref{fig:pipeline}.
We use DPDP in the speech-to-unit encoder of the SLM, varying the codebook size $K$ and the coarseness of the units through the parameter $\lambda$ in~\eqref{eq:dpdp_objective}.
A unit language model is trained to predict the next unit.
For tasks where we want to generate audio, a unit-to-speech decoder maps the units back to speech.
Each of these components is now described in detail.

\textbf{Speech-to-unit (S2U) encoder:}
Our S2U encoder (Figure~\ref{fig:pipeline}-top) starts by extracting features from layer~11 of WavLM Large~\cite{chen2021wavlm}.
We chose this layer because it resulted in the best phone discriminability in development experiments.
We train \mbox{$K$-means} codebooks with sizes $K \in \{100, 200, 500, 1000\}$.
Each codebook is trained on 10\% of the features from Libri\-Speech \texttt{train-clean-100}~\cite{panayotov2015librispeech}.
We use the \texttt{k-means++} initialization from \texttt{scikit-learn} and train for 300 iterations using the Faiss~library~\cite{douze2024faiss}.

We then apply DPDP quantization.
We vary $\lambda$ for each codebook size to get six bitrate values, ranging from the full bitrate of the original $K$-means units to half that bitrate (as in Figure~\ref{fig:dp_units}).
As a secondary contribution of this work, we update the DPDP algorithm so that, instead of searching the full codebook at each timestep, it only searches for solutions where the resulting codebook entry is within the nearest 5\% of neighbors to the feature.
This dramatically speeds up the inference time without affecting performance (tested in development experiments).

\textbf{Unit language model (ULM):} 
Our ULM (Figure~\ref{fig:pipeline}-middle) uses the transformer implementation from Mistral~\cite{jiang2023mistral7b} to predict the next unit in the sequence.
The model has 200M non-embedding parameters.
We use an embedding dimension of 1024, 12 layers, 16 attention heads, a hidden dimension of 4096, and rotary positional embeddings~\cite{su2021roformer}.
The ULM is trained with a maximum context window size of 1024, holding a contiguous chunk of units from a single chapter in Libri\-Speech.
We train a different ULM for each combination of codebook size $K$ and coarseness value $\lambda$.
Each model is trained on the full 960~hours of Libri\-Speech for 10k steps using a batch size of 100k tokens.

\textbf{Unit-to-speech (U2S) decoder:}
For some tasks, we want the SLM to be generative and include a unit-to-speech (U2S) decoder, i.e., a pseudo-text-to-speech model.
Our U2S decoder uses an acoustic model to add duration, speaker, and acoustic information and predicts the acoustic codes from WavTokenizer~\cite{ji2024wavtokenizer}.
WavTokenizer is a neural audio codec trained to compress acoustic information into a stream of discrete codes with a codebook size of 4096.
The pre-trained WavTokenizer decoder allows for high-fidelity synthesis from these codes.

To predict the WavTokenizer acoustic codes from the varying-length DPDP units (the output from S2U encoder), we use ELLA-V \cite{song2024ellav}.
Figure~\ref{fig:pipeline} (bottom) illustrates how ELLA-V tokenizes its training data by sandwiching the relevant acoustic codes (blue circles) between a unit~(e.g., $\otimes$) and end-of-unit~(\texttt{EOU}) token.
ELLA-V then uses an autoregressive transformer model to predict the acoustic tokens.
The interleaving strategy helps the transformer to remain aware of the current unit for which it needs to predict acoustic codes.
Therefore, \mbox{ELLA-V} does not suffer from generating repetitions or long silences, unlike VALL-E~\cite{wang2023valle}.
ELLA-V is well-suited to work with larger varying-length units, even up to syllable size \cite{baade2024syllablelm}.

We train the ELLA-V acoustic model on 460 hours of clean audio from Libri\-Speech.
The autoregressive transformer used within ELLA-V has the same structure and capacity as the one used in the ULM.
Each context window includes the global advance and interleaved tokens for one utterance.
We train with a batch size of 64 utterances.
We retrain the combined U2S decoder for each combination of $K$ and $\lambda$.

\section{Experiments}

We want to know how codebook size and unit coarseness (how long the units are) affect SLM performance.
Do coarser representations aid language modeling?
How large can we make the units while still being able to obtain the original content from them, in intelligible synthesized speech?
To answer these questions, we conduct evaluations across different linguistic levels, ranging from discriminative tasks at the phone and word level to tasks that consider whole sentences.

In Figures~\ref{fig:results_abx_samediff_vs_bitrate} through \ref{fig:results_swuggy_sblimp}, we present a metric against the bitrate of the units.
Each curve represents a single codebook size $K$.
The point at the highest bitrate (marked in black) corresponds to directly using $K$-means and deduplicating the units.
Sets of coarser units are obtained by increasing the DPDP parameter $\lambda$, which decreases the bitrate (moving left on each plot).

\subsection{Phone and word discrimination}
\label{sec:abx_samediff}

\hspace{\parindent} \textbf{ABX task:} The ABX task \cite{schatz2013abx} evaluates how well the discretized speech representations can discriminate between phone classes.
The test asks whether a phone segment $x$ is more similar to phone segments $a$ or $b$.
Here, $a$ and $x$ would be instances of the same phone class while $b$ would differ.
The phone segments are all drawn from encoded codebook sequences.
We specifically use the any-context within-speaker ABX variant \cite{hallap23_interspeech}.

\begin{figure}[!t]
  \centering
  \includegraphics[width=0.99\linewidth]{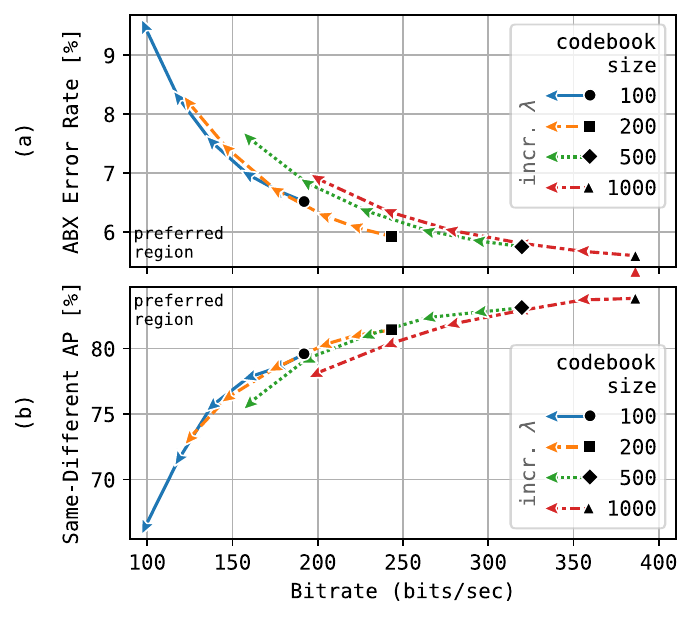}
  \vspace*{-10pt}
  \caption{Phone and word discrimination scores of the DPDP units (colored curves) compared to \mbox{$K$-means} units (black markers). (a)~ABX error rates. (b)~Same-different average precision (AP). Both tasks are evaluated on LibriSpeech~\texttt{dev-clean}.}
  \label{fig:results_abx_samediff_vs_bitrate}
  \vspace{-17pt}
\end{figure}

\textbf{Results:} 
Figure~\ref{fig:results_abx_samediff_vs_bitrate}~(a) shows the ABX error rate against bitrate. 
It is favorable to be in the lower left corner, i.e., an ABX error of zero with as few bits as possible.
The ABX error rate increases substantially when DPDP is applied to obtain coarser units at lower bitrates.
Systems using DPDP units offer no advantage over systems using $K$-means units.
For example, when operating at a fixed bit rate, say 190~bits/sec, the ABX error rate is lowest when we use the $K$-means units with $K=100$.
Other systems that use DPDP to get coarser units from $K>100$ achieve either the same or higher error rates.
Thus, for phone discrimination, coarser units prove detrimental to performance.
The coarser units are context-dependent and lack the fine phonetic distinctions required for ABX.\footnote{In Figure \ref{fig:results_abx_samediff_vs_bitrate}, we used the any-context ABX variant, but we did find a coarser representation to help in the within-context ABX task. This explains how we know that the units here are context-dependent.}

\textbf{Same-different task:}
At a higher linguistic level, the same-different task \cite{carlin2011samedifferent} evaluates how well the discretized features perform in identifying whether two spoken segments correspond to the same word.
Given pairs of word segments $w_i$ and $w_j$, the task uses the minimum dynamic time warping alignment cost to classify the words as the same if $\mathrm{DTW}(w_i,w_j) \leq \tau$, where $\tau$ is some threshold.
By sweeping $\tau$ and constructing a precision-recall curve, we compute the average precision~(AP).

\textbf{Results:}
Figure~\ref{fig:results_abx_samediff_vs_bitrate} (b) shows word discrimination performance against bitrate.
It is favorable to be in the upper-left corner.
A similar conclusion can be drawn from the same-different task compared to the ABX task: DPDP units offer no performance-per-bit advantage over the fine-grained $K$-means units.

\subsection{Resynthesis}

\hspace{\parindent} \textbf{Task:}
Although coarser units do not allow for discriminating between isolated phone or word units, it could be that the units still capture this information if longer contexts are taken into account.
To test this, we see if we can resynthesize intelligible speech from the coarse DPDP units.
Since we are only interested in whether the content of the original message is preserved without mispronunciation, we do not care about the output voice.
We, therefore, do not condition the ELLA-V model on any speaker information during generation.
We use a greedy approach when sampling from the ELLA-V model, which reduces expressivity but gives more stable outputs for these comparative experiments.
After synthesis, we transcribe the speech with the HuBERT-Large ASR model~\cite{hsu2021hubert} and compare the output to ground truth transcriptions.
A lower word error rate (WER) indicates that content is better preserved.

\begin{figure}[!t]
  \centering
  \includegraphics[width=0.99\linewidth]{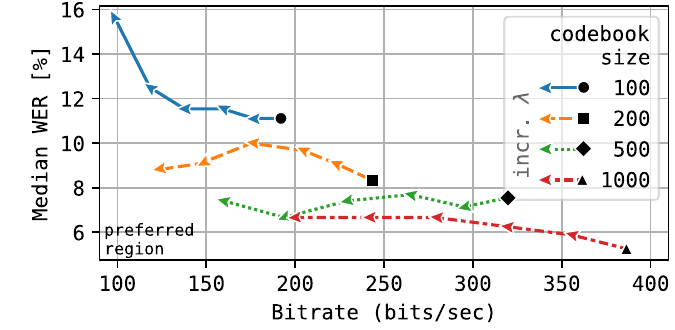}
  \vspace*{-7.5pt}
  \caption{WERs when resynthesizing the \texttt{dev-clean} subset of LibriSpeech using increasingly coarser DPDP units (colored curves) compared to the $K$-means units (black markers).}
  \label{fig:results_resynthesis}
  \vspace{-15pt}
\end{figure}

\textbf{Resynthesis results:}
Figure~\ref{fig:results_resynthesis} shows the median WER when transcribing full sentences that were resynthesized from the units.
It is favorable to be in the lower-left corner of Figure~\ref{fig:results_resynthesis}.
For a small codebook size of $K=100$, the coarser units (solid blue) cause a high amount of mispronunciation in the resynthesized output.
However, coarser units obtained from larger codebooks do not introduce a substantial number of mispronunciations, and the WER remains relatively stable even at low bitrates.
If we fix the bitrate at 190 bits/sec, there is a definite advantage in using DPDP units.
The coarser units obtained from a larger codebook size $K=500$ result in a much lower WER compared to using the $K$-means units with~$K=100$.
While the larger context-dependent units are detrimental to phone discrimination, they are beneficial to speech synthesis -- provided that the codebook size is large enough to support a variety of contexts.

\subsection{Language modeling}

To further see how coarseness affects performance at higher linguistic levels, we evaluate SLM performance using the lexical and syntactic tasks of the Zero Resource Speech Challenge~\cite{dunbar2021zrc}.

\textbf{Lexical task (sWUGGY):}
In the sWUGGY task, the model is presented with pairs of spoken words: a real word like ``manufacture" and a matched non-word like ``manufelture."
The task is to score each word and identify the real word from the non-word using log-likelihood scores from the ULM.

\begin{figure}[ht]
  \centering
  \includegraphics[width=0.99\linewidth]{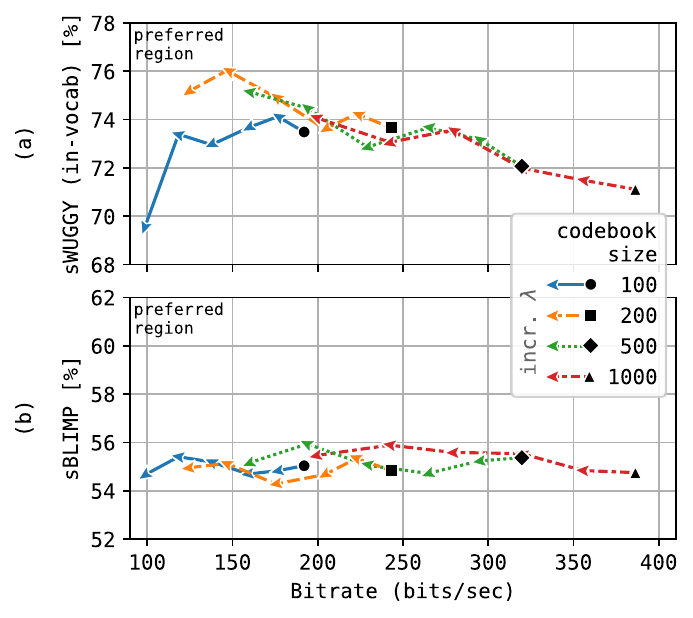}
  \vspace*{-10pt}
  \caption{Language modeling performance of the DPDP units (colored curves) compared to the $K$-means units (black markers). (a)~sWUGGY lexical accuracy. (b)~sBLIMP syntactic accuracy. Both tasks are evaluated on sLM21 \texttt{dev}~\cite{dunbar2021zrc}.}
  \label{fig:results_swuggy_sblimp}
  \vspace{-5pt}
\end{figure}

\textbf{Results:}
Figure~\ref{fig:results_swuggy_sblimp} (a) shows how coarser units affect the lexical task.
Here we want to be in the upper-left corner.
When working with a small codebook size of $K=100$, the sWUGGY accuracy drops as we use coarser units (left in plot).
However, for all larger codebook sizes, the sWUGGY accuracy increases as we use coarser units.
Therefore, using coarser units obtained through DPDP benefits language modeling at the lexical level if the codebook size is chosen appropriately.

\textbf{Syntactic task (sBLIMP):}
In the sBLIMP task, the model receives pairs of sentences: one is syntactically correct, and the other has a minimal modification to give it incorrect syntax.
The SLM must score and identify the correct sentence.
For example: ``What \textit{dentist} is Angela working with?" vs.\ ``What is Angela working with \textit{dentist}?"

\textbf{Results:}
While the results in Figure~\ref{fig:results_swuggy_sblimp} (b) are not as clear-cut as in previous tasks, there are
DPDP systems that provide better per-bit performance compared to the $K$-means systems.
The best overall results are achieved with a DPDP-based system: $K = 500$ (green dotted curve) at 190 bits/sec.
This system reduces sequence lengths by 40\%.

\textbf{Computational benefit:}
Since we train with a fixed context length, language models using coarser units train faster.
The number of steps required reduces proportionally with the sequence length for all the systems considered.

\subsection{Comparison to state-of-the-art}

Our goal is a scientific investigation looking into the effect of codebook size and unit coarseness on an SLM, and not an attempt to achieve state-of-the-art results.
Nevertheless, we briefly compare to other systems using a scaled version of our SLM.
We use a codebook size of 500 at 190 bits/sec, scale the ULM training data to 60k hours of speech from Libri-Light~\cite{librilight2020}, and train the model for 200k steps.
We did not try any other settings.
We compare this model to other SLMs in Table~\ref{tbl:lm_results_compare}.
When scaled, our DP-SLM model approaches the scores of AudioLM~\cite{borsos2023audiolm} and SyllableLM \cite{baade2024syllablelm}.
Although we do not outperform these systems, our method is more memory efficient than AudioLM and does not require training an intricate model like SyllableLM.
Future work will consider explicitly optimizing the hyperparameters and architectures of DP-SLM for state-of-the-art results.

\begin{table}[!t]
    \centering
    \caption{Language modeling performance compared to other state-of-the-art SLMs on the sLM21 \texttt{dev} benchmark~\cite{dunbar2021zrc}.}
    \vspace*{-5pt}
    \label{tbl:lm_results_compare}
    \renewcommand{\arraystretch}{0.8}
    \begin{tabularx}{1.0\linewidth}{@{}l@{\ }RRCcc@{}}
        \toprule
         & {Bitrate} & {Data} & \multicolumn{2}{c}{{sWUGGY}} & {sBLIMP} \\ 
         \cmidrule(lr){4-5}
         & bps& hours & all & in-vocab & \\
        \midrule
        Phone Topline                         &  53 &  1k &  78.3 & 92.2 & 63.0 \\
        \midrule
        GSLM~\cite{lakhotia2021gslm}          & 332 &  6k &    - & 68.7 & 57.1 \\
        AudioLM~\cite{borsos2023audiolm}      & 250 & 60k & 71.5 & \textbf{83.7} & \textbf{64.7} \\
        SyllableLM~\cite{baade2024syllablelm} & \textbf{81} & 60k & \textbf{72.2} & 82.2 & 63.7 \\
        \midrule
        DP-SLM [ours]                         & 190 &  1k & 66.0 & 74.5 & 56.0 \\
        DP-SLM [ours]                         & 190 & 60k & 70.8 & 81.1 & 60.7 \\
        \bottomrule
    \end{tabularx}
    \vspace{-10pt}
\end{table}

\section{Discussion}

How do codebook size and unit coarseness affect spoken language models (SLMs), particularly when using duration-penalized dynamic programming (DPDP) for unit extraction?
We investigated the interaction between these two crucial aspects in a range of evaluations moving up the linguistic hierarchy.
Taking the results together, it appears that coarser units are not beneficial at the lowest levels in discriminating between isolated units like phones or words.
But as we move up the hierarchy, we saw gains in resynthesis and lexical language modeling tasks when using coarser DPDP units.
Our results also establish DPDP as a simple and efficient method to obtain coarser units in contexts where shorter input lengths are crucial.

In our final experiments, we looked at state-of-the-art SLMs trained on up to 60k hours of speech audio.
A text-based model trained on only 1k hours of phone sequences still outperforms all of the speech models.
So SLMs still have a long way to go -- this gap (at 1k hours) will be the focus of our future work.

\section{Acknowledgements}

This research was partly supported by the Harry Crossley Foundation and the Google PhD fellowship program.

\bibliographystyle{IEEEtran}
\bibliography{mybib}

\end{document}